\documentclass[10pt,twocolumn,letterpaper]{article}

\usepackage[pagenumbers]{cvpr} 

\usepackage[accsupp]{axessibility}  

\usepackage{booktabs}
\usepackage{array}
\usepackage{xcolor}
\usepackage{colortbl}
\usepackage{graphicx}
\usepackage{makecell}
\usepackage{multirow}
\usepackage{pifont}
\usepackage{amssymb}
\usepackage{bm}
\usepackage{tabularx}
\usepackage{float}

\definecolor{lightgreen}{RGB}{144,238,144}

\definecolor{cvprblue}{rgb}{0.21,0.49,0.74}
\usepackage[pagebackref,breaklinks,colorlinks,allcolors=cvprblue]{hyperref}

\def\paperID{} 
\def\confName{CVPR}
\def\confYear{2026}

\title{GeoPredict: Leveraging Predictive Kinematics and 3D Gaussian Geometry for Precise VLA Manipulation}

\author{
    Jingjing Qian$^{1}$ \quad
    Boyao Han$^{2}$ \quad
    Chen Shi$^{1}$ \quad
    Lei Xiao$^{1}$ \quad
    Long Yang$^{1}$ \quad
    Shaoshuai Shi$^{3}$ \quad
    Li Jiang$^{1\dagger}$ \\
    $^{1}$The Chinese University of Hong Kong, Shenzhen \\
    $^{2}$Hunan University \quad
    $^{3}$Voyager Research, Didi Chuxing \\[5pt]
    {\tt\small \url{https://jingjingqian75.github.io/GeoPredict-Page/}}
}

\begin{document}
\maketitle
{
    \renewcommand{\thefootnote}{$\dagger$}
    \footnotetext{Corresponding Author.}
}
\begin{abstract}
Vision--Language--Action (VLA) models achieve strong generalization in robotic manipulation but remain largely reactive and 2D-centric, making them unreliable in tasks that require precise 3D reasoning. We propose \textbf{GeoPredict}, a geometry-aware VLA framework that augments a continuous-action policy with predictive kinematic and geometric priors. GeoPredict introduces a trajectory-level module that encodes motion history and predicts multi-step 3D keypoint trajectories of robot arms, and a predictive 3D Gaussian geometry module that forecasts workspace geometry with track-guided refinement along future keypoint trajectories. These predictive modules serve exclusively as training-time supervision through depth-based rendering, while inference requires only lightweight additional query tokens without invoking any 3D decoding. Experiments on RoboCasa Human-50, LIBERO, and real-world manipulation tasks show that GeoPredict consistently outperforms strong VLA baselines, especially in geometry-intensive and spatially demanding scenarios.

\end{abstract}    
\section{Introduction}
\label{sec:intro}

Vision--Language--Action (VLA) models have recently emerged as a powerful paradigm for robotic manipulation, leveraging large pre-trained Vision--Language Models (VLMs) to map rich visual observations and language instructions directly to actions~\cite{brohan2022rt, zitkovich2023rt, li2024cogact, zheng2025universal}. By inheriting semantic and visual priors from internet-scale data, these models generalize well across tasks and settings. However, they operate primarily in 2D image space, mapping the current observation reactively and lacking explicit 3D spatial modeling~\cite{zhong2025survey}. This 2D-centric, myopic formulation limits performance in tasks requiring precise 3D reasoning and physically consistent long-horizon control.

To move beyond purely reactive policies, recent work has begun to incorporate predictive structure into visuomotor models~\cite{hu2025video, li2025unified, lu2025gwm}. Some approaches learn latent dynamics or temporal abstractions to better capture how states evolve over time~\cite{chen2025villa, chen2025moto, li2024vision, zhao2025cot}, while others predict future observations such as RGB frames, depths, or point-based representations~\cite{cen2025worldvla,zhangdreamvla,hu2025video,qian2025wristworld, grauman2022ego4d,gupta2024pre}. Although these methods provide useful temporal signals, they often remain view-independent and do not strictly enforce multi-view or 3D geometric consistency, making it hard to reason about object poses, clearances, and end-effector motion in workspace coordinates. Moreover, tightly coupling high-capacity predictive modules with large VLA backbones can introduce non-negligible computational overhead if complex 3D prediction is required during inference time.

For robotic manipulation, we argue that two predictive capabilities are particularly valuable. First, the policy should exploit \emph{predictive kinematic priors} that summarize how the robot is likely to move over the next few steps, rather than relying solely on instantaneous joint states. Second, it should reason about \emph{predictive 3D Gaussian geometry} through an explicit, differentiable representation aligned with the robot’s workspace and amenable to supervision from depth and multi-view cues. Crucially, these predictive signals must seamlessly integrate with VLA architectures and keep inference-time overhead modest enough to enable practical deployment for real-time robotic control.

In this paper, we introduce \textbf{GeoPredict}, a geometry-aware VLA framework augmented with predictive priors for robotic manipulation. GeoPredict is built on top of a strong continuous-action VLA policy~\cite{black2410pi0}, and augments the policy with two complementary predictive modules.

First, at the kinematic level, we introduce a \emph{trajectory-level prediction module} that tracks key robot joints and end-effector points and encodes their motion history into compact tokens via a Track Encoder. A set of learnable future track queries then predicts multi-step 3D keypoint trajectories, providing explicit, differentiable kinematic priors. These predicted trajectories not only regularize the transformer's internal dynamics, but also serve as guidance for where high-fidelity 3D scene modeling is most needed.

Second, at the geometric level, we introduce a \emph{predictive 3D Gaussian geometry module} that forecasts how workspace geometry evolves over time. A coarse 3D spatial query grid is decoded into Gaussian primitives representing the scene at future timesteps. To focus geometric capacity where precision matters most, a track-guided refinement mechanism increases Gaussian density along the predicted future keypoint trajectories of the joints, yielding higher resolution around anticipated motion and interaction regions. The resulting 3DGS representation is supervised through future depth-map rendering, providing a geometry-focused training signal without requiring color modeling.

Both predictive modules are integrated via a block-wise causal attention hierarchy and are used only during training; inference behaves exactly like the base VLA policy.

In summary, our main contributions are threefold:
(1) We introduce \textbf{GeoPredict}, a geometry-aware VLA framework that injects future-aware kinematic and geometric priors into a continuous-action VLA policy, enhancing reasoning about long-horizon 3D dynamics.
(2) We propose two complementary predictive modules: a trajectory-level kinematic predictor forecasting multi-step robot keypoint motion, and a predictive 3D Gaussian geometry module with track-guided refinement that allocates geometric capacity to task-relevant interaction regions.
(3) We show that GeoPredict delivers consistent and substantial improvements over strong VLA baselines on RoboCasa Human-50, LIBERO, and real-world manipulation tasks, particularly those requiring precise spatial reasoning and geometric robustness.

\section{Related Work} \label{sec:related_work}

\subsection{Vision-Language-Action Models}

Vision-Language-Action (VLA) models have emerged as powerful robotic policies~\cite{brohan2022rt, kim2025openvla, li2024cogact, zheng2024tracevla, zheng2025universal, xiao2025ava}. Pioneering methods like OpenVLA~\cite{kim2025openvla} rely on discrete autoregressive tokenization, limiting inference frequency and struggling with continuous, multi-modal action distributions~\cite{zhong2025survey}. While subsequent architectures like $\pi_0$~\cite{black2410pi0} utilize conditional flow matching~\cite{lipman2022flow} to generate continuous actions, they remain predominantly 2D-centric and reactive~\cite{bu2025univla, black2410pi0, hu2025video, zheng2024tracevla}, lacking the robust 3D geometric understanding required for complex manipulation. Furthermore, although SpatialVLA~\cite{qu2025spatialvla} and BridgeVLA~\cite{li2025bridgevla} integrate explicit 3D information and auxiliary tasks, they still lack a predictive understanding of 3D scene dynamics. To address this gap, our GeoPredict introduces a geometry-aware predictive model that directly forecasts 3D scene evolution using 3D Gaussian Splatting~\cite{kerbl20233d}, equipping the VLA model with predictive geometric priors.

\subsection{Future Prediction for Robotic Manipulation}

Recent works explore future prediction to enhance robotic policies~\cite{cen2025worldvla, zhang2023storm, hafnermastering, micheli2022transformers, hansen2023td, hafner2022deep}. Some directly forecast future observations like RGB images~\cite{wu2024unleashing, hu2025video}, depths~\cite{zhangdreamvla}, or pointmaps~\cite{liu2025geometry}, while approaches like Seer~\cite{tianpredictive} condition inverse dynamics on these forecasted visual states. However, these methods face two key limitations. First, popular approaches like SuSiE~\cite{black2023zero} or UniPi~\cite{du2023learning} often rely on a single prediction step, which struggles to capture complex physical dynamics and introduces time-consuming denoising that reduces control frequency. Second, these 2D-centric methods~\cite{wu2024unleashing} struggle with multi-view consistency and lack the explicit 3D geometry crucial for accurate manipulation. To address this gap, we introduce GeoPredict, a geometry-aware predictive model that forecasts scene evolution over a full horizon of $H$ steps in explicit 3D space, learning a spatially coherent and predictive 3D representation to condition action generations.

\subsection{3D Scene Representation} 

3D scene representation is a fundamental problem in computer vision, traditionally relying on explicit structures like meshes~\cite{lorensen1998marching}, voxels~\cite{wu20153d} and point clouds~\cite{qi2017pointnet}. Recent research has shifted to neural representations that learn continuous models of scene geometry and appearance. Pioneered by Neural Radiance Fields (NeRF)~\cite{mildenhall2021nerf}, these implicit models often require subsequent acceleration~\cite{muller2022instant,xu2022point,hu2022efficientnerf} due to costly training and rendering. More recently, 3D Gaussian Splatting (3DGS)~\cite{kerbl20233d,yan2024street,wu20244d,ren2024scube} has emerged as a highly effective explicit alternative. 3DGS models scenes as a collection of 3D Gaussians with learned attributes, leveraging differentiable rasterization for high-fidelity and remarkably efficient rendering. While recent works like GWM~\cite{lu2025gwm} apply 3DGS to robotic world modeling, directly forecasting these massive Gaussian attributes remains computationally demanding. To circumvent this burden, we leverage 3DGS as a predictive scene geometry module solely during training, aiming to endow the VLA policy with intrinsic, foresight-driven spatial reasoning ability.

\section{Methodology}
\label{sec:method}

\begin{figure*}[htbp]
  \centering
  \includegraphics[width=0.99\textwidth]{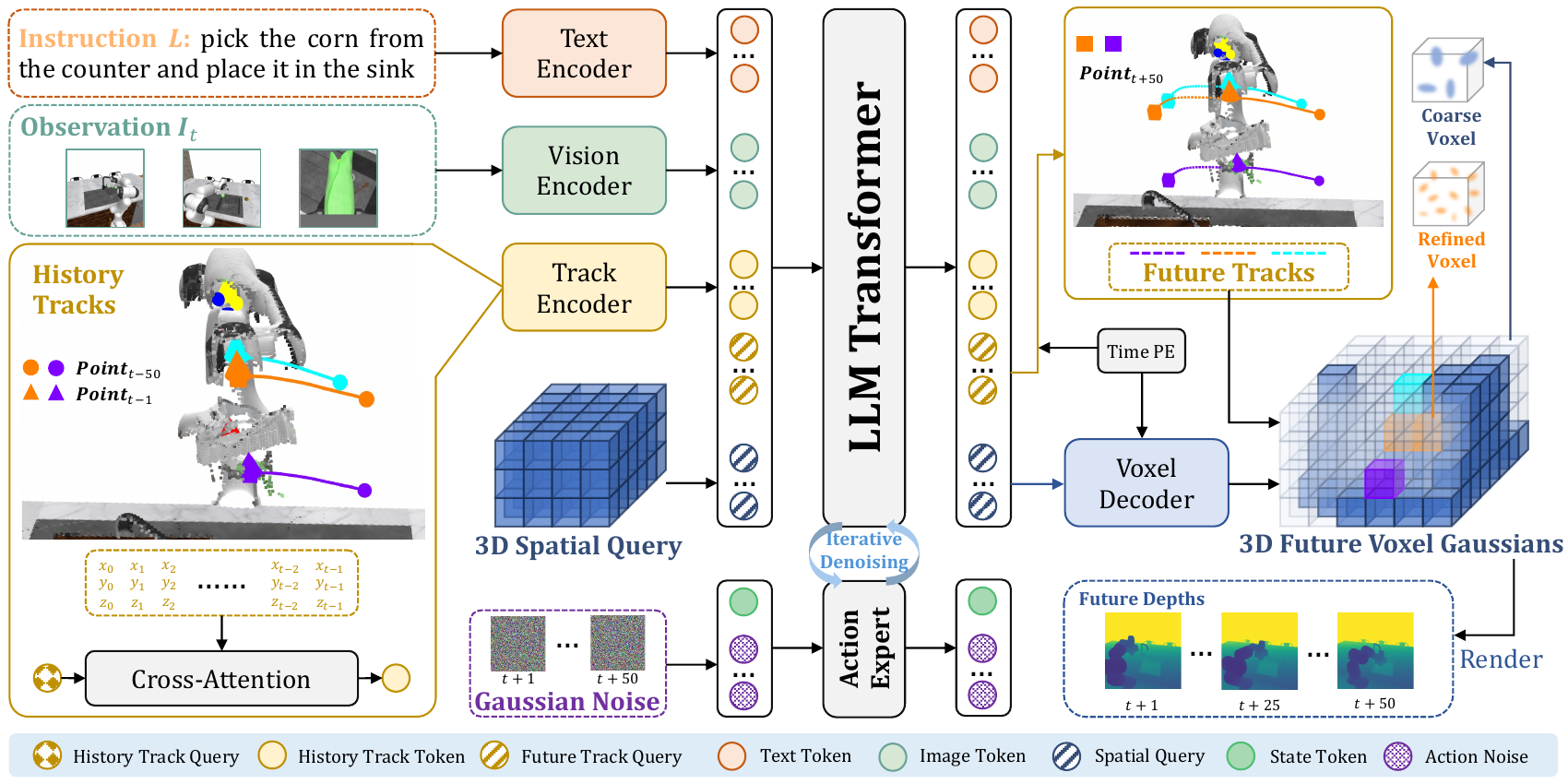}
  \caption{\textbf{Overview of GeoPredict.} Given an instruction, multi-view images and motion history encoded by the Track Encoder, a central LLM Transformer learns two main tasks. First, it predicts multi-timestep 3D keypoint trajectories using learnable Future Track Query. Second, it forecasts future workspace geometry as a predictive 3D Gaussian by processing a 3D Spatial Query through a Voxel Decoder. A track-guided refinement mechanism leverages the predicted future tracks to allocate geometric capacity to task-relevant interaction regions. Our policy then generates the final action via an Action Expert. Crucially, these predictive modules serve exclusively as training-time supervision and are not invoked during inference, thus preserving efficiency.}
\label{fig:method}
\vspace{-15pt}
\end{figure*}

We propose \textbf{GeoPredict}, a geometry-aware VLA framework that equips continuous-control policies with predictive kinematic and geometric priors. GeoPredict jointly learns future robot keypoint trajectories and future scene geometry through a 3DGS representation, and conditions the underlying VLA on these predictions. Notably, these predictive modules are used only during training, while the action generation process remains similar to a standard VLA policy.

First, we describe the problem setup and the base VLA model (Sec.~\ref{sec:preliminary}). We then introduce our trajectory encoding and future motion prediction module (Sec.~\ref{sec:track_query}), followed by the predictive 3D Gaussian geometry and its refinement mechanism (Sec.~\ref{sec:3dgs}). Finally, we present the attention design, training objective and inference process (Sec.~\ref{sec:train_infer}).

\subsection{Preliminary} \label{sec:preliminary}

\textbf{Problem Definition.}
%
%
The VLA paradigm~\cite{kim2025openvla,black2410pi0} aims to learn a policy $\bm{\pi}$ that maps a language instruction $\mathbf{L}$, the current multi-view image observation $\mathbf{I}_t$, and the robot’s proprioceptive state $\mathbf{Q}_t$ to a continuous action chunk $\mathbf{A}_t=[\mathbf{a}_t, \mathbf{a}_{t+1}, \ldots, \mathbf{a}_{t+H-1}]$, where we set $H=50$. 
Each action $\mathbf{a}_t \in \mathbf{R}^{7}$ is a 7-DoF end-effector command $\mathbf{a}_t=\{\Delta\mathbf{x}, \Delta\bm{\theta}, g\}$,
where $\Delta\mathbf{x}\in\mathbb{R}^3$ and $\Delta\bm{\theta}\in\mathbb{R}^3$ denote translational and rotational offsets (e.g., in axis–angle or Euler representation), and $g\in\mathbb{R}$ is the gripper's open-close state.


While  VLAs generate actions reactively from the current observation, many manipulation tasks require anticipating how both the robot and the surrounding scene will evolve over time. Thus, GeoPredict augments the policy by conditioning not only on past observations but also on internally predicted future robot motion and future scene geometry.

%
%

\textbf{Base VLA model.}
We adopt $\pi_0$~\cite{black2410pi0} as the VLA base model. It comprises a VLM (PaliGemma~\cite{beyer2024paligemma} with a SigLIP vision encoder) and a robotics-specific \emph{action expert} that generates continuous actions via conditional flow matching~\cite{lipman2022flow}. Text, image, and proprioceptive tokens are projected into a shared space and processed by the transformer; the action expert then generates $\mathbf{A}_t$ by integrating a learned vector field from noise in a small, fixed number of steps.



\subsection{Trajectory-Level Kinematic Prediction} \label{sec:track_query}


GeoPredict models robot motion at the trajectory level. This provides both a compact summary of past kinematics and an explicit prediction of future trajectories, serving as a kinematic prior for action generation and as guidance for spatial refinement in the following 3D Gaussian geometry module.

\noindent \textbf{Track Encoder.}
%
%
The Track Encoder compresses the motion history of all robot keypoints into compact tokens for the transformer. This historical context provides a strong kinematic prior: because joint motion exhibits inertia and action generation fundamentally involves forecasting future trajectories, encoding past dynamics enables the model to make more physically consistent motion predictions.

As illustrated in Figure~\ref{fig:method}, we track $K$ 3D keypoints (joints and end-effector points). For each keypoint $k\in\{1,\ldots,K\}$, we collect its 3D coordinates from time $0$ to $t{-}1$ into a trajectory $\bm{\mathcal{T}}_k \in \mathbb{R}^{(t-1)\times 3}$. A shared learnable history query $\mathbf{Q}^{\text{hist}}$ performs cross-attention over the embedded trajectory to produce a single \textit{history track token}:
\vspace{-10pt}
\begin{equation}
\begin{aligned}
\mathbf{Z}_k^{\text{hist}}
= \text{CrossAttn}\big(&\text{query}=\mathbf{Q}^{\text{hist}},\;
\text{key}=\text{MLP}(\bm{\mathcal{T}}_k),\\
&\text{value}=\text{MLP}(\bm{\mathcal{T}}_k)\big).
\end{aligned}
\end{equation}
These $K$ history tokens encode inertia, joint limits, and motion regularities, and are appended to the transformer input.

\noindent \textbf{Future Track Query.}
%
%
To explicitly capture future motion, we introduce $K$ learnable future track queries $\{\mathbf{q}_k^{\text{fut}}\}_{k=1}^K$. These queries are processed together with the instruction $\mathbf{L}$, current images $\mathbf{I}_t$, and history tokens $\{\mathbf{Z}_k^{\text{hist}}\}$, producing future track embeddings $\mathbf{e}_k^{\text{fut}}$ that summarize the latent predicted trajectory of each keypoint.


We decode each $\mathbf{e}_k^{\text{fut}}$ into an explicit spatio-temporal trajectory by using a shared MLP to predict $H{+}1$ timesteps, one for the current time $t$ and $H$ for future steps $\{t{+}1,\ldots,t{+}H\}$. A 1D sinusoidal temporal encoding $\mathbf{PE}^{\text{time}}\in\mathbb{R}^{(H+1)\times C}$ is then adopted as follows:
\begin{equation}
\label{eq:track}
\hat{\mathbf{p}}_{k,t+\tau}
= \text{MLP}\big(\mathbf{e}_k^{\text{fut}} + \mathbf{PE}^{\text{time}}[\tau]\big),
\qquad \tau = 0,\ldots,H.
\end{equation}

The predicted trajectories are supervised using a Mean Squared Error loss over all keypoints and timesteps as 
\begin{equation}
\label{eq:track_loss}
\mathcal{L}_{\text{track}}
= \frac{1}{K(H+1)}
\sum_{k=1}^{K}\sum_{\tau=0}^{H}
\|\hat{\mathbf{p}}_{k,t+\tau} - \mathbf{p}^{\text{gt}}_{k,t+\tau}\|_2^2.
\end{equation}
This encourages the shared transformer backbone to learn predictive, dynamically consistent motion representations and provides explicit future trajectories that will be used for predictive 3D Gaussian refinement.

\subsection{Predictive 3D Gaussian Geometry} \label{sec:3dgs}


We now introduce a predictive 3D Gaussian geometry module that estimates the future geometry of the robot workspace. By anticipating how the scene will evolve, this module provides the policy with stronger spatial reasoning abilities, which is essential for manipulation tasks that reply on forecasting future object configurations. The module comprises four components: a 3D spatial query for encoding the robot workspace, a voxel decoder that produces 3D Gaussian Splatting (3DGS) primitives, a track-guided refinement mechanism for increasing fidelity near interaction regions, and a depth-rendering objective for supervision.

\smallskip
\noindent
\textbf{3D Spatial Query.}
%
%
To represent the workspace efficiently, we define a 3D volume of dimensions $H \times W \times D$ meters and discretize it into voxels of size $v$, yielding a grid resolution of $(H/v) \times (W/v) \times (D/v)$. For token efficiency, this high-resolution grid is downsampled by a factor of 4 along each axis, producing a coarse grid of size $N_x \times N_y \times N_z$, where $N_x = H/(4v)$, $N_y = W/(4v)$, and $N_z = D/(4v)$.


Each coarse voxel is assigned a learnable $C$-dimensional embedding, forming the initial spatial query tensor $\mathbf{Q}^{\text{init}} \in \mathbb{R}^{N_x \times N_y \times N_z \times C}$. To incorporate geometric structure and reduce learning difficulty, we add a 3D sinusoidal positional encoding $\mathbf{PE}^{\text{spatial}}\in \mathbb{R}^{N_x \times N_y \times N_z \times C}$, constructed by concatenating independent 1D sinusoidal encodings along the $x$, $y$, and $z$ axes:
$
\mathbf{PE}^{\text{spatial}}[i,j,k]
= \text{Concat}\big(\mathbf{PE}^\text{x}[i],\; \mathbf{PE}^\text{y}[j],\; \mathbf{PE}^\text{z}[k]\big).
$
Hence, the final spatial query is obtained by
\begin{equation}
\mathbf{Q}^{\text{spatial}}[i,j,k]
= \mathbf{Q}^{\text{init}}[i,j,k] + \mathbf{PE}^{\text{spatial}}[i, j,k].
\end{equation}
All spatial queries are flattened into a sequence of  $N_x N_y  N_z$  tokens and provided to the transformer together with the instruction, current observations and history track tokens.



\smallskip
\noindent \textbf{Voxel Decoder.}
After multi-layer attention, we obtain spatial embeddings $\mathbf{E}^{\text{spatial}} \in \mathbb{R}^{(N_x N_y N_z) \times C}$ that summarize the workspace representation at time $t$. To model future geometry, we reuse the temporal positional encoding $\mathbf{PE}^{\text{time}}$ from Eq.~\eqref{eq:track} and construct temporally shifted embeddings
\begin{equation}
\mathbf{E}^{\text{spatial}}_{t+\tau} = \mathbf{E}^{\text{spatial}} + \mathbf{PE}^{\text{time}}[\tau], \quad \tau=0,\ldots,H.
\end{equation}
Each $\mathbf{E}^{\text{spatial}}_{t+\tau}$ is reshaped into a grid of size $N_x \times N_y \times N_z$ and passed through a 3D voxel decoder composed of transposed convolutions and upsampling layers, which restores the original voxel resolution $(H/v) \times (W/v) \times (D/v)$. This yields a dense voxel feature volume
\begin{equation}
\mathbf{F}^{\text{voxel}} \in \mathbb{R}^{(H/v) \times (W/v) \times (D/v) \times C'},
\end{equation}
where $C'$ is the feature dimension of the voxel feature grid. 


A final 3D convolution maps each voxel feature $\mathbf{F}^{\text{voxel}}[i,j,k]$ to a set of $N_G$ Gaussian primitives. Following the 3DGS formulation~\cite{kerbl20233d}, each primitive
$
\mathbf{g}=\{\bm{\mu},\,\alpha,\,\bm{\Sigma}\}
$
has a 3D center $\bm{\mu}\in\mathbb{R}^3$, an opacity $\alpha\in\mathbb{R}$, and a covariance matrix $\bm{\Sigma}\in\mathbb{R}^{3\times 3}$. Because our focus is on geometry rather than appearance, we omit color coefficients. The union of Gaussians over all voxels forms the initial 3D scene representation $\mathbf{G}^{\text{init}}_{t+\tau}$.

\smallskip
\noindent \textbf{Track-based Gaussian Refinement.}
%
While a coarse global representation is often sufficient for capturing the overall workspace, accurate manipulation requires high-fidelity geometry near regions of interaction, particularly around the end-effector, joints, and task-relevant objects. We therefore refine the Gaussian representation adaptively using the predicted future keypoint trajectories from Sec.~\ref{sec:track_query}.


At timestep $t+\tau$, given the initial Gaussian set $\mathbf{G}^{\text{init}}_{t+\tau}$ and the predicted keypoint positions $\mathbf{P}_{t+\tau}=\{\hat{\mathbf{p}}_{k,t+\tau}\}_{k=1}^{K}$, we define a binary refinement mask for each voxel $\bm{\mathcal{V}}[i,j,k]$:
\begin{equation}
\small
\mathbf{M}^{\text{refine}}[i,j,k]=
\begin{cases}
1, & \text{if } \exists\, \mathbf{p}\in \mathbf{P}_{t+\tau} ~~\text{s.t.}~~ \mathbf{p}\in \bm{\mathcal{V}}[i,j,k], \\
0, & \text{otherwise}.
\end{cases}
\end{equation}




For each voxel with $\mathbf{M}^{\text{refine}}[i,j,k]=1$, a shared MLP maps $\mathbf{F}^{\text{voxel}}[i,j,k]$ to an additional $N_G'$ finer-grained Gaussian primitives ($N_G' > N_G$), yielding a refined set $\mathbf{G}^{\text{refine}}_{t+\tau}$. The complete Gaussian representation for timestep $t+\tau$ is 
\vspace{-3pt}
\begin{equation}
\mathbf{G}^{\text{total}}_{t+\tau}
= \mathbf{G}^{\text{init}}_{t+\tau} \cup \mathbf{G}^{\text{refine}}_{t+\tau}.
\vspace{-3pt}
\end{equation}


This track-guided refinement concentrates modeling capacity near predicted interaction regions, improving geometric accuracy along the robot's anticipated motion path without requiring a globally high-resolution scene model.

\noindent 
\textbf{Future Depth Rendering.}
%
To supervise the predictive 3DGS model, we render depth maps from $\mathbf{G}^{\text{total}}_{t+\tau}$ at all $H{+}1$ timesteps using differentiable alpha compositing, following the rendering formulation of 3D Gaussian Splatting~\cite{kerbl20233d}. For a given camera and timestep $t{+}\tau$, we consider a pixel ray $\mathbf{r}$ and collect all Gaussians whose projected ellipses intersect the ray into a set $\mathcal{N}$, sorted in front-to-back order. For the $i$-th Gaussian, let $d_i$ denote the depth of its center $\bm{\mu}_i$ and $\alpha_i$ its opacity. The accumulated transmittance is
\vspace{-5pt}
\begin{equation}
T_i \;=\; \prod_{j=1}^{i-1} (1 - \alpha_j),
\vspace{-5pt}
\end{equation}
representing the probability that the ray passes through all primitives in front of the $i$-th one. The rendered depth for ray $\mathbf{r}$ is then
\vspace{-5pt}
\begin{equation}
\vspace{-5pt}
\hat{\mathbf{D}}(\mathbf{r}) \;=\; \sum_{i\in\mathcal{N}} T_i\,\alpha_i\, d_i.
\end{equation}
Since we focus on geometry prediction, colors are omitted.

To restrict supervision to regions relevant for manipulation, we apply a spatial mask over the robot’s operational workspace. For each ray $\mathbf{r}$, we back-project the depth value to obtain a 3D point $\mathbf{p}_{3d}(\mathbf{r})$. If this point lies within the predefined workspace, we set $\mathbf{M}^{\text{spatial}}(\mathbf{r})=1$; otherwise, $\mathbf{M}^{\text{spatial}}(\mathbf{r})=0$. The depth loss is formulated as a masked L1 loss over all cameras and timesteps:
\vspace{-5pt}
\begin{equation}
\vspace{-5pt}
\label{eq:depth_loss}
\begin{split}
\mathcal{L}_{\text{depth}}
&= \frac{1}{\sum \mathbf{M}^{\text{spatial}}}
\sum_{\tau=0}^{H} \sum_{c=1}^{N_{cam}} \sum_{\mathbf{r} \in \text{pixels}}\\
&\mathbf{M}^{\text{spatial}}(\mathbf{r})\,
\big| \hat{\mathbf{D}}_{c,t+\tau}(\mathbf{r})
- \mathbf{D}^{\text{gt}}_{c,t+\tau}(\mathbf{r}) \big|.
\end{split}
\end{equation}
This loss encourages the predicted 3D Gaussian representation to accurately capture the evolution of workspace geometry in regions relevant to downstream manipulation.


\subsection{Training and Inference} \label{sec:train_infer}

We now introduce how the kinematic and geometric prediction modules are integrated into the transformer via a block-wise causal attention mechanism, and how the full system is jointly trained and deployed efficiently.

\begin{figure}[t]
  \centering
  \includegraphics[width=0.47\textwidth]{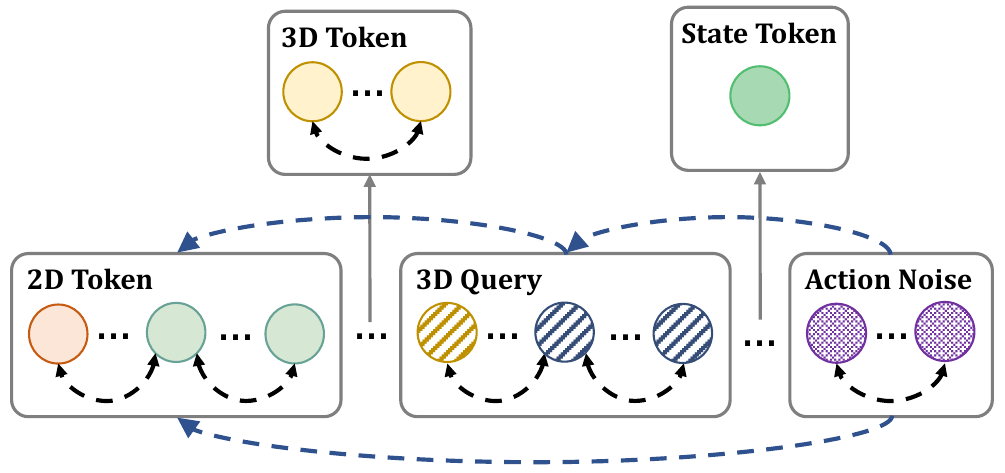}
  \caption{\textbf{Block-wise Causal Attention Mechanism.} For simplicity, the detailed attention pathways from the 3D Token and State Token blocks to other blocks are not fully drawn.}
  \label{fig:attention}
  \vspace{-10pt}
\end{figure}

\noindent
\textbf{Block-wise Causal Attention Mechanism.}
%
%
We adopt a block-wise causal attention structure adapted from $\pi_0$ as shown in Figure~\ref{fig:attention}, in which the full token sequence is partitioned into five conceptually ordered groups: (1) 2D Token (text and image tokens), (2) 3D Token (history track tokens), (3) 3D Query tokens (future track queries and spatial queries), (4) State Token (proprioceptive token) and (5) Action Noise tokens (used by flow matching). Attention is fully bidirectional within each block, enabling rich intra-block interactions, while cross-block attention is strictly causal: tokens in a given block may attend only to tokens within the same block or to those in preceding blocks.



This organization imposes a perceptual-to-predictive-to-control hierarchy. The transformer fuses 2D language and multi-view observations, incorporates 3D kinematic history, forms predictive 3D representations (future trajectories and spatial fields), integrates current proprioception, and generates an action via the action-noise block. The predictive modules thus occupy an intermediate stage, shaping control by injecting structured motion and geometry priors into the transformer.


\noindent
\textbf{Training Objective.}
%
We train the model end-to-end using a weighted sum of three loss terms:
\vspace{-5pt}
\begin{equation}
    \mathcal{L}_{\text{total}}
    = \lambda_{1}\mathcal{L}_{\text{action}}
    + \lambda_{2}\mathcal{L}_{\text{track}}
    + \lambda_{3}\mathcal{L}_{\text{depth}},
    \label{eq:total_loss}
    \vspace{-5pt}
\end{equation}
where $\mathcal{L}_{\text{action}}$ is the continuous conditional flow-matching loss from $\pi_0$, $\mathcal{L}_{\text{track}}$ is the trajectory prediction loss defined in Eq.~\ref{eq:track_loss}, and $\mathcal{L}_{\text{depth}}$ is the masked depth-rendering loss from Eq.~\ref{eq:depth_loss}. The coefficients $\lambda_{1}$, $\lambda_{2}$, and $\lambda_{3}$ balance supervision from action, kinematic, and geometric signals.


\noindent
\textbf{Inference.}
%
At inference time, all context tokens including text, images, history track tokens and 3D queries are processed by the transformer in a single forward pass, which computes and caches attention Key--Value pairs for subsequent use. The \emph{action expert} then performs its iterative denoising process for $a$ steps to generate the action chunk $\mathbf{A}_t$, reusing the cached Key--Value states and invoking only its lightweight action-specific layers.

Importantly, the predictive 3D Gaussian geometry modules (e.g., the voxel decoder and depth-rendering module) are not executed during inference. They serve purely as training-time supervision signals that shape the transformer's internal representations with geometric and physical structure, while the action-generation pathway remains unchanged and efficient during deployment.


\section{Experiments} \label{sec:experiments}






We first introduce our experimental setup (Sec.~\ref{sec:exp_setup}), followed by a benchmarking against state-of-the-art baselines (Sec.~\ref{sec:main_results}). We then dissect the contribution of our proposed modules through detailed ablation studies (Sec.~\ref{sec:ablations}) and qualitative analysis (Sec.~\ref{sec:qualitative}). Finally, we assess the model's performance in real-world scenarios (Sec.~\ref{sec:real_world}).

\subsection{Experimental Setup} \label{sec:exp_setup}

\noindent \textbf{Simulation Benchmarks.}
We evaluate our method on two widely recognized robotic manipulation benchmarks.
\begin{itemize}[nolistsep, topsep=0pt, itemsep=1pt, parsep=0pt]
    \item \textbf{RoboCasa}~\cite{nasiriany2024robocasa}: This benchmark features 24 complex long-horizon everyday tasks in kitchen environments. We adopt the challenging \texttt{Human-50} few-shot setting, where policies are trained using only 50 human demonstrations per task. To rigorously assess generalization capabilities, we follow the standard protocol by evaluating our model over 50 trials per task across five distinct scenes. This evaluation is conducted exclusively on \textit{unseen object instances}, and includes scenes with \textit{unseen styles} that were never encountered during training.
    
    \item \textbf{LIBERO}~\cite{liu2023libero}: This benchmark is designed to evaluate knowledge transfer and policy generalization across four distinct task suites: LIBERO-Spatial, LIBERO-Object, LIBERO-Goal, and LIBERO-Long. We follow the standard protocol for LIBERO, training policies on 50 human-teleoperated demonstrations for each task and evaluating over 50 trials per task (500 trials per suite).
\end{itemize}

\begin{figure}[t]
  \centering
  \includegraphics[width=0.47\textwidth]{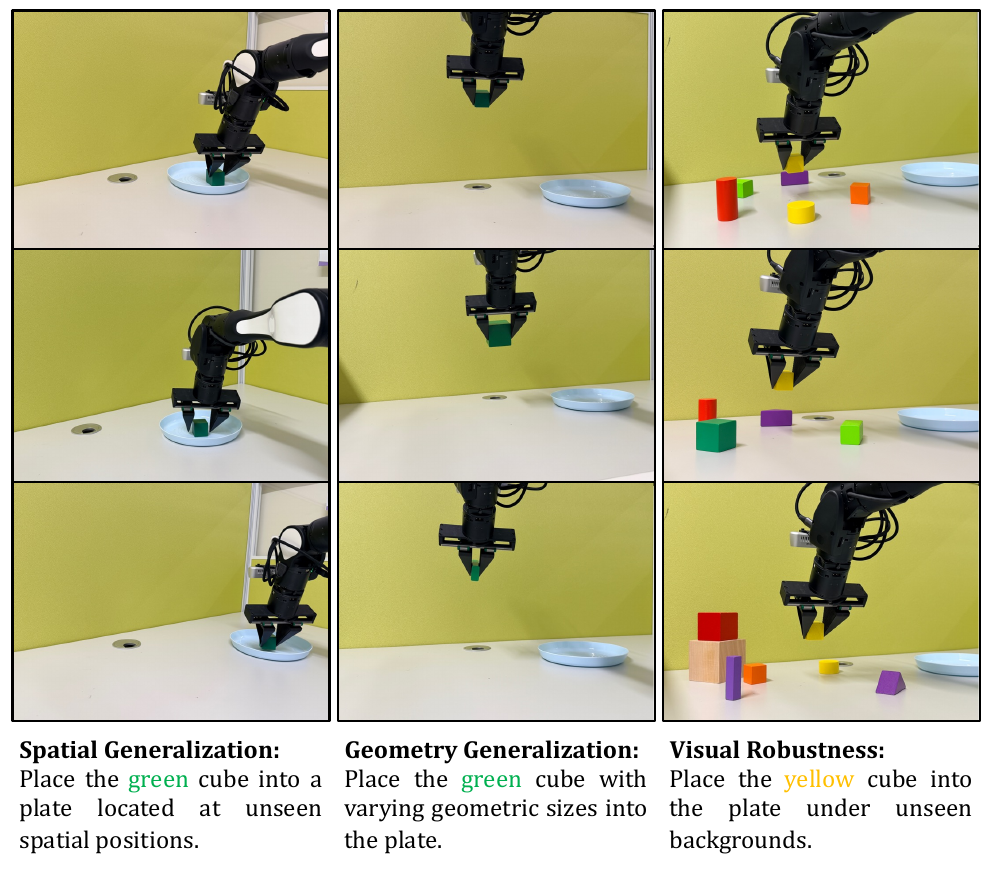}
  \vspace{-5pt}
  \caption{\textbf{Real-world Evaluation Suite.} These settings aim to evaluate the model's spatial generalization, geometry generalization and robustness to distractors. Each column represents different trials of the same task.}
  \label{fig:real}
  \vspace{-10pt}
\end{figure}

\noindent \textbf{Real-world Evaluation Suite.}
To evaluate GeoPredict's efficacy on real-world tasks requiring precise 3D reasoning, we use an evaluation suite (Figure~\ref{fig:real}). This suite includes three task categories, using 50 expert trajectories per category for training. Each policy is evaluated for 20 trials per category, where success is defined as successfully grasping the target object and placing it in the correct position.

\begin{itemize}[nolistsep, topsep=0pt, itemsep=1pt, parsep=0pt]
    \item \textbf{Spatial Generalization:} The robot is instructed to place the green cube into a plate. We evaluate the policy with the plate located at positions \textit{unseen} during training, testing its ability to generalize to novel target locations.

    \item \textbf{Geometry Generalization:} During the 20 evaluation trials, the model must grasp four distinct object types: a small cube, a medium cube, a large cube, and a rectangular prism. Critically, the test set includes cubes with dimensions \textit{not seen} during training. Furthermore, the rectangular prism is tested in its three unique stable orientations, requiring the policy to adapt its grasp approach based on the perceived object geometry.

    \item \textbf{Visual Robustness:} This task evaluates the policy's resilience to visual distractors. The robot must place the yellow cube into the plate while novel and task-irrelevant objects are introduced into the background of the scene, which were absent during training.
\end{itemize}

\begin{table*}[htbp]
\caption{\textbf{RoboCasa Simulation Benchmark Results.} Task success rates (\%) across 24 sub-tasks and the Average Success Rate (\%). $^{\bm{\ast}}$Denotes our fine-tuned experimental results. \textbf{Bold} indicates the best performing model. See Appendix for detailed sub-task definitions.}
\vspace{-5pt}
\centering
\small
\begin{tabular}{l|cccccccccccc} 
\toprule
\multirow{2}{*}{\textbf{Method}} & \textbf{PnP} & \textbf{PnP} & \textbf{PnP} & \textbf{PnP} & \textbf{PnP} & \textbf{PnP} & \textbf{PnP} & \textbf{PnP} & \textbf{CM} & \textbf{CM} & \textbf{CB} & \multirow{2}{*}{\textbf{TSS}} \\
 & \textbf{CTC1} & \textbf{CTC2} & \textbf{CTS1} & \textbf{STC1} & \textbf{CTS2} & \textbf{STC2} & \textbf{CTM} & \textbf{MTC} & \textbf{SU} & S\textbf{V} & \textbf{PS} & \\ 
\midrule

BC-Transformer~\cite{nasiriany2024robocasa}& 6.0 & 2.0 & 2.0 & 8.0 & 2.0 & 6.0 & 2.0 & 2.0 & 0.0 & 22.0 & 48.0 & 54.0 \\

GWM~\cite{lu2025gwm} & 4.0 & \textbf{18.0} & 20.0 & 22.0 & 2.0 & 18.0 & 14.0 & \textbf{20.0} & 16.0 & 36.0 & \textbf{76.0} & \textbf{72.0} \\

$\pi_0^{\bm{\ast}}$ (Baseline)~\cite{black2410pi0} & 24.0 & 6.8 & 26.8 & 15.6 & 11.2 & 12.8 & 6.8 & 7.6 & 18.4 & 33.6 & 74.0 & 69.6 \\

\textbf{GeoPredict (Ours)} & \textbf{32.4} & 8.8 & \textbf{27.6} & \textbf{31.2} & \textbf{20.4} & \textbf{28.8} & \textbf{14.4} & 18.0 & \textbf{28.4} & \textbf{49.2} & 67.6 & 70.8 \\

\midrule
\textbf{Average Success Rate} & 
\textbf{OSD} &
\textbf{CSD} &
\textbf{ODD} &
\textbf{CDD} &
\textbf{OD} &
\textbf{CD} &
\textbf{TNSF} &
\textbf{TFSF} &
\textbf{TNS} &
\textbf{TFS} &
\textbf{TNM} &
\textbf{TFM} \\
\midrule

\multicolumn{1}{c|}{28.8} & 46.0 & 56.0 & 28.0 & 28.0 & 42.0 & 80.0 & 38.0 & 50.0 & 32.0 & 4.0 & 62.0 & 70.0\\

\multicolumn{1}{c|}{39.2} & \textbf{58.0} & 54.0 & 28.0 & 50.0 & 56.0 & 80.0 & 52.0 & 44.0 & 46.0 & \textbf{22.0} & 64.0 & 70.0\\

\multicolumn{1}{c|}{42.3} & 40.8 & \textbf{82.0} & 55.2 & 69.2 & 66.4 & 96.0 & 43.6 & 86.0 & 43.2 & 6.0 & 59.6 & 60.0\\

\multicolumn{1}{c|}{\textbf{52.4}} & 40.4 & 78.8 & \textbf{87.2} & \textbf{70.0} & \textbf{77.6} & \textbf{96.8} & \textbf{72.4} & \textbf{94.8} & \textbf{60.0} & 13.2 & \textbf{84.8} & \textbf{82.8}\\

\bottomrule
\end{tabular}
\label{tab:robocasa}
\vspace{-5pt}
\end{table*}

\begin{table*}[htbp]
\centering
\small
\caption{\textbf{LIBERO Simulation Benchmark Results.} Task success rates (\%) across 4 evaluation suites and the Average Success Rate (\%). $^{\bm{\ast}}$Denotes our reproduced experimental results. $^{\bm{\dagger}}$Denotes no available standard deviation data. \textbf{Bold} indicates the \textbf{best} performing model, and \underline{underline} indicates the \textbf{runner-up} model.}
\vspace{-5pt}
\begin{tabular}{l | c  c  c  c | c}
\toprule
\textbf{Method} & 
\multicolumn{1}{c}{\textbf{LIBERO-Spatial}} & 
\multicolumn{1}{c}{\textbf{LIBERO-Object}} & 
\multicolumn{1}{c}{\textbf{LIBERO-Goal}} & 
\multicolumn{1}{c|}{\textbf{LIBERO-Long}} & 
\multicolumn{1}{c}{\textbf{Average}} \\
\midrule
Diffusion Policy~\cite{chi2025diffusion} & 78.3 $\pm$ 1.1 & 92.5 $\pm$ 0.7 & 68.3 $\pm$ 1.2 & 50.5 $\pm$ 1.3 & 72.4 $\pm$ 0.7 \\
TraceVLA~\cite{zheng2024tracevla} & 84.6 $\pm$ 0.2 & 85.2 $\pm$ 0.4 & 75.1 $\pm$ 0.3 & 54.1 $\pm$ 1.0 & 74.8 $\pm$ 0.5 \\
Octo~\cite{team2024octo} & 78.9 $\pm$ 1.0 & 85.7 $\pm$ 0.9 & 84.6 $\pm$ 0.9 & 51.1 $\pm$ 1.3 & 75.1 $\pm$ 0.6 \\
OpenVLA~\cite{kim2025openvla} & 84.7 $\pm$ 0.9 & 88.4 $\pm$ 0.8 & 79.2 $\pm$ 1.0 & 53.7 $\pm$ 1.3 & 76.5 $\pm$ 0.6 \\
SpatialVLA~\cite{qu2025spatialvla} & 88.2 $\pm$ 0.5 & 89.9 $\pm$ 0.7 & 78.6 $\pm$ 0.6 & 55.5 $\pm$ 1.0 & 78.1 $\pm$ 0.7 \\
WorldVLA$^{\bm{\dagger}}$~\cite{cen2025worldvla} & 87.6 & 96.2 & 83.4 & 60.0 & 81.8 \\
4D-VLA~\cite{zhang20254d} & 88.9 $\pm$ 0.5 & 95.2 $\pm$ 0.3 & 90.9 $\pm$ 0.4 & 79.1 $\pm$ 1.2 & 88.6 $\pm$ 0.3 \\
DreamVLA$^{\bm{\dagger}}$~\cite{zhangdreamvla} & \underline{97.5} & 94.0 & 89.5 & 89.5 & 92.6 \\
$\pi_0^{\bm{\dagger}}$~\cite{black2410pi0} & 96.8 & \textbf{98.8} & \textbf{95.8} & 85.2 & 94.2\\
OpenVLA-OFT$^{\bm{\dagger}}$~\cite{kim2025fine} & 95.2 & 94.2 & 95.2 & \underline{93.2} & 94.5 \\
UniVLA$^{\bm{\dagger}}$~\cite{bu2025univla} & 96.5 & 96.8 & 95.6 & 92.0 & \underline{95.2} \\
\midrule
$\pi_0^{\bm{\ast}}$ (Baseline)~\cite{black2410pi0} & 96.6 $\pm$ 0.6 & 97.2 $\pm$ 0.8 & 94.2 $\pm$ 0.7 & 87.6 $\pm$ 1.1 & 93.9 $\pm$ 0.4\\
\textbf{GeoPredict (Ours)} & \textbf{98.0 $\pm$ 0.7} & \underline{98.2 $\pm$ 0.7} & \underline{95.7 $\pm$ 0.2} & \textbf{94.0 $\pm$ 1.0} & \textbf{96.5 $\pm$ 0.6} \\
\bottomrule
\end{tabular}
\label{tab:libero}
\vspace{-10pt}
\end{table*}

\noindent \textbf{Baselines.}
Our primary baseline is our VLA backbone, \textbf{$\pi_0$}~\cite{black2410pi0}, trained without our proposed predictive 3D modules. This comparison directly isolates the contribution of our geometry-aware predictive method. We also compare against a range of state-of-the-art VLA and world-modeling approaches, including BC-Transformer~\cite{nasiriany2024robocasa}, GWM~\cite{lu2025gwm}, OpenVLA~\cite{kim2025openvla}, SpatialVLA~\cite{qu2025spatialvla} and UniVLA~\cite{bu2025univla}.

\noindent \textbf{Implementation Details.}
A predictive horizon of $H=50$ is used for all tasks. Observations include two environment and one wrist camera views. Depth supervision is applied only to the two $224\times224$ environment cameras. For the trajectory-level prediction module (Sec.~\ref{sec:track_query}), we track $K=8$ keypoints (7 joints, 1 end-effector) for LIBERO~\cite{liu2023libero} and RoboCasa~\cite{nasiriany2024robocasa}, and $K=7$ (6 joints, 1 end-effector) for the real-world setup. For the predictive 3DGS geometry module (Sec.~\ref{sec:3dgs}), the workspace is $1.6\text{m} \times 1.6\text{m} \times 1.0\text{m}$ with $v=0.04$m voxels. Token and decoder feature dimensions are $C=2048$ and $C'=256$. We use $N_G=4$ initial primitives per voxel, and generate $N_G'=64$ additional primitives for voxels selected by the track-guided refinement mechanism. All loss weights in Eq.~\eqref{eq:total_loss} are 1.0. We train for 40,000 iterations using AdamW~\cite{loshchilovdecoupled} (LR 2.5e-5) on 8 NVIDIA H20 GPUs with a total batch size of 32. The primary metric is Task Success Rate (\%).

\subsection{Main Results} \label{sec:main_results}

\noindent \textbf{RoboCasa.}
Table~\ref{tab:robocasa} details the performance across the 24 RoboCasa sub-tasks. GeoPredict achieves an average success rate of \textbf{52.4\%}, significantly outperforming all baselines. Notably, it improves upon the $\pi_0$ baseline success rate of 42.3\% by a margin of \textbf{10.1\%}. This substantial gain underscores the benefit of conditioning the policy on predictive kinematic and geometric priors, especially in the \texttt{Human-50} few-shot setting where generalization from limited data is critical~\cite{nasiriany2024robocasa}. Moreover, GeoPredict demonstrates clear advantages over other future prediction methods such as GWM~\cite{lu2025gwm} at 39.2\% and 2D policies like BC-Transformer~\cite{nasiriany2024robocasa} at 28.8\%. Additional large-scale results on RoboCasa \texttt{Generated-300} are in the Appendix.

\noindent \textbf{LIBERO.}
In Table~\ref{tab:libero}, we show results on the four LIBERO suites~\cite{liu2023libero}. GeoPredict achieves an average success rate of \textbf{96.5\%}, surpassing the current SOTA method UniVLA~\cite{bu2025univla}. Furthermore, it significantly outperforms strong baselines such as OpenVLA~\cite{kim2025openvla} which attains 76.5\% by a margin of 20.0\%. Our model demonstrates consistently high performance across all four suites, including Spatial (98.0\%), Object (98.2\%), Goal (95.7\%), and the challenging Long (94.0\%). This indicates that our method's geometry-aware training is highly effective and generalizes robustly across diverse tasks and environments.

\subsection{Ablation Study} \label{sec:ablations}
\noindent \textbf{Component Ablation.}
Table~\ref{tab:ab_module} details our ablation, building the model from the ground up. Starting from the $\pi_0$ baseline (42.3\%), adding the history Track Encoder provides a kinematic prior, improving performance to 44.8\%. Adding the Future Track Query ($\mathcal{L}_{track}$) for explicit motion prediction further boosts the rate to 47.2\%.

The predictive 3D Gaussian Splatting geometry module yields the most substantial gains. Supervising with $\mathcal{L}_{depth}$ (using only initial $\mathbf{G}^{init}$) elevates performance to 49.4\%. Jointly training with $\mathcal{L}_{track}$ and $\mathcal{L}_{depth}$ but without the explicit \textit{track-guided refinement} yields 50.5\%. Critically, activating the full \textit{track-guided refinement} mechanism, which allocates geometric capacity along the predicted future tracks, achieves the peak 52.4\%. This final 1.9\% gain (50.5\% $\to$ 52.4\%) validates our hypothesis: using kinematic predictions to refine the 3DGS representation provides a superior geometric prior for the policy.

\noindent \textbf{Ablation on Depth Rendering.}
Table~\ref{tab:ab_depth} ablates our predictive 3DGS method. Rendering with color (49.2\%) offers no benefit over depth-only (49.4\%), confirming geometric information alone suffices. Increasing the global initial primitives from $N_G=4$ to $N_G=8$ (51.4\%) significantly increases training time (12.0 to 19.1 hours/epoch).

In contrast, our \textit{track-guided refinement} strategy ($N_G=4, N_G'=8$) is more efficient, achieving 51.1\% at only 15.5 hours/epoch. This highlights the efficiency of allocating geometric capacity to task-relevant interaction regions. Since these regions are a small fraction of the volume, we can increase $N_G'$ with negligible overhead. Setting $N_G'=64$ yields our peak 52.4\% success rate at a near-identical 15.7 hours. Therefore, we adopt $N_G=4$ and $N_G'=64$ as our final configuration.

\begin{table}[t]
\centering
\small
\caption{\textbf{Ablation Study on RoboCasa Simulation Benchmark.} Future Depth is rendered from initial global Gaussians ($\textbf{G}^{init}$), while Future Depth$^{\bm{\ast}}$ utilizes the refined total Gaussians ($\textbf{G}^{total}$). Results denote the Average Success Rate (\%) across all sub-tasks.}
\vspace{-5pt}
\begin{tabular}{l | c  c  c  c c c}
\toprule
History Track & \textcolor{red}{\ding{56}} & \textcolor{blue}{\ding{52}} & \textcolor{blue}{\ding{52}} & \textcolor{red}{\ding{56}} & \textcolor{blue}{\ding{52}} & \textcolor{blue}{\ding{52}}\\
Future Track & \textcolor{red}{\ding{56}} & \textcolor{red}{\ding{56}} & \textcolor{blue}{\ding{52}} & \textcolor{red}{\ding{56}} & \textcolor{blue}{\ding{52}} & \textcolor{blue}{\ding{52}}\\
Future Depth & \textcolor{red}{\ding{56}} & \textcolor{red}{\ding{56}} & \textcolor{red}{\ding{56}} & \textcolor{blue}{\ding{52}} & \textcolor{blue}{\ding{52}} & \textcolor{red}{\ding{56}}\\
Future Depth$^{\bm{\ast}}$ & \textcolor{red}{\ding{56}} & \textcolor{red}{\ding{56}} & \textcolor{red}{\ding{56}} & \textcolor{red}{\ding{56}} & \textcolor{red}{\ding{56}} & \textcolor{blue}{\ding{52}}\\
\midrule
\textbf{Average SR} & 42.3 & 44.8 & 47.2 & 49.4 & 50.5 & 52.4\\
\bottomrule
\end{tabular}
\label{tab:ab_module}
\vspace{-3pt}
\end{table}

\begin{table}[t]
\centering
\small
\caption{\textbf{Ablation Study on Depth Rendering.} $N_G$ and $N_G'$ denote the number of Gaussian primitives per voxel for the initial and refined stages, respectively. Color indicates RGB image reconstruction. \textbf{Time / Epoch} represents the training time per epoch in hours, and \textbf{Average SR} is the Average Success Rate (\%).}
\vspace{-5pt}
\begin{tabular}{l | c  c  c  c c c}
\toprule
$N_G$ & 4 & 4 & 8 & 4 & 4 \\
$N_G'$ & \textcolor{red}{\ding{56}} & \textcolor{red}{\ding{56}} & \textcolor{red}{\ding{56}} & 8 & 64 \\
Color & \textcolor{blue}{\ding{52}} & \textcolor{red}{\ding{56}} & \textcolor{red}{\ding{56}} & \textcolor{red}{\ding{56}} & \textcolor{red}{\ding{56}} \\
\midrule
\textbf{Time / Epoch} & 12.3 & 12.0 & 19.1 & 15.5 & 15.7 \\
\textbf{Average SR} & 49.2 & 49.4 & 51.4 & 51.1 & 52.4\\
\bottomrule
\end{tabular}
\label{tab:ab_depth}
\vspace{-3pt}
\end{table}

\subsection{Qualitative Analysis} \label{sec:qualitative}
\begin{figure}[t]
  \centering
  \includegraphics[width=0.47\textwidth]{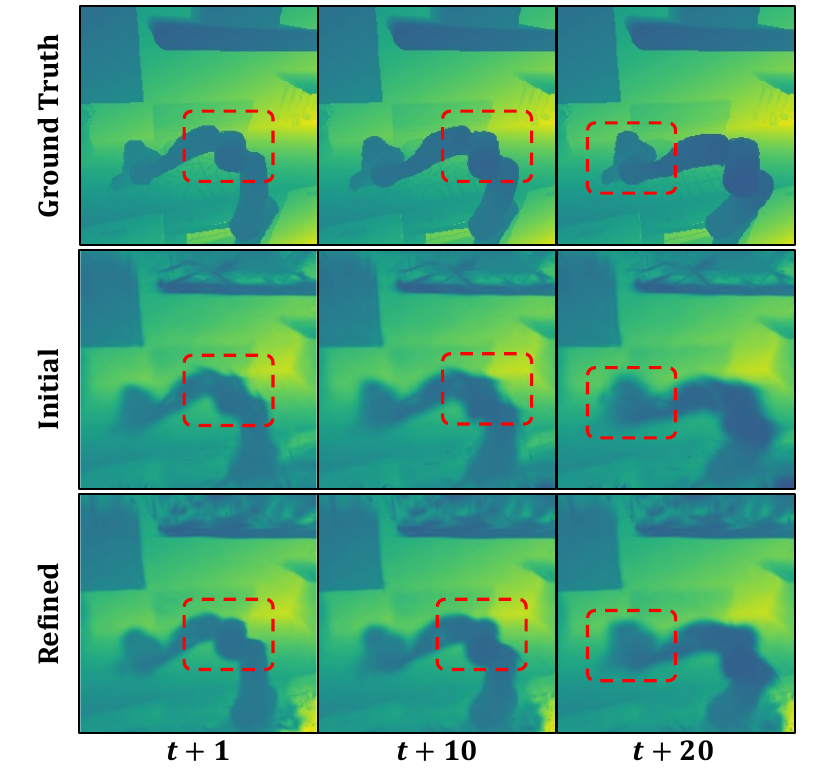}
  \vspace{-5pt}
  \caption{\textbf{Qualitative Comparisons of Future Depth Rendering.} Visualizations are shown for timesteps $t+1, t+10, \text{and } t+20$. \textcolor{red}{Red boxes} highlight the improvements in fine-grained geometric details. Best viewed zoomed in.}
  \vspace{-10pt}
  \label{fig:vis}
\end{figure}

Figure~\ref{fig:vis} provides a qualitative visualization of our predictive 3DGS geometry module, which compares the predicted future depths at various timesteps ($t+1, t+10, t+20$). While the \textbf{initial} Gaussians ($\mathbf{G}^{init}$) capture only the coarse scene layout, the \textbf{refined} Gaussians ($\mathbf{G}^{total}$) exhibit significantly sharper geometric details, particularly surrounding the robotic arm. This visually confirms that our refinement mechanism produces a more precise and geometrically-faithful prediction of the future, which in turn provides a superior conditional prior for the action expert.

\begin{table}[t]
\centering
\small
\caption{\textbf{Real-World Experiment Results.} Task success rates (\%) across three distinct settings: Spatial, Geometry, and Robustness.}
\vspace{-5pt}
\begin{tabular}{l | c c c}
\toprule
\textbf{Method} & \textbf{Spatial} & \textbf{Geometry} & \textbf{Robustness}\\
\midrule
$\pi_0$ (Baseline)~\cite{black2410pi0} & 60.0 & 50.0 & 35.0  \\
\textbf{GeoPredict (Ours)} & 85.0 & 95.0 & 90.0 \\
\bottomrule
\end{tabular}
\label{tab:real}
\vspace{-7pt}
\end{table}

\subsection{Real-World Experiments} \label{sec:real_world}
We validate GeoPredict on a DISCOVER robotic arm to assess performance on tasks demanding explicit 3D reasoning. As detailed in Table~\ref{tab:real}, GeoPredict significantly outperforms the $\pi_0$~\cite{black2410pi0} baseline across all tasks. Specifically, in the Spatial and Robustness tasks, our method achieves success rates of 85.0\% and 90.0\% respectively, surpassing the baseline scores of 60.0\% and 35.0\% and demonstrating superior spatial generalization and resilience to distractors. In the Geometry task, GeoPredict attains a 95.0\% success rate compared to 50.0\% for the baseline. This 45.0\% improvement on tasks involving object sizes absent from the training set strongly indicates that our predictive 3DGS module endows the policy with a generalizable understanding of 3D geometry, enabling adaptive grasping capabilities that the 2D-centric $\pi_0$ baseline lacks. Moreover, since our method explicitly encodes motion history via the Track Encoder, the Appendix provides a dedicated real-world experiment to validate the effectiveness of this historical context.

\section{Conclusion}
\label{sec:conclusion}

We presented \textbf{GeoPredict}, a geometry-aware VLA framework enhancing continuous-action policies with predictive kinematic and geometric priors. By forecasting robot trajectories and workspace geometry via track-guided 3DGS, GeoPredict introduces future-aware structure into the transformer while keeping inference lightweight. Extensive experiments show consistent superiority over strong baselines in tasks requiring precise 3D reasoning, geometric robustness and long-horizon physical consistency. While acquiring multi-view RGB-D and camera extrinsics presents scaling challenges, the increasing availability of calibrated depth in modern datasets and commodity hardware mitigates these concerns. Ultimately, our training-time predictive supervision provides a promising direction for reliable, geometry-aware, and grounded VLA controllers.

\section*{Acknowledgment}
{\sloppy
This work is supported by the Shenzhen Science and Technology Program (Grant No. ZDCY20250901113000001) and the 1+1+1 CUHK-CUHK(SZ)-GDSTC Joint Collaboration Fund (Grant No. 2025A0505000079). We gratefully acknowledge DISCOVER Robotics for providing hardware support.
\par}

{
    \small
    \bibliographystyle{ieeenat_fullname}
    \bibliography{main}
}


\end{document}